\documentclass{ifacconf}
\usepackage{eurosym}
\usepackage{float,adjustbox}
\usepackage{caption}
\usepackage{natbib}
\usepackage{amsmath,amssymb}
\usepackage{subfigure}
\usepackage{graphicx}
\usepackage{subfloat}
\usepackage{float}

\begin{document}

\begin{frontmatter}

\title{Model- and Data-Based Control of Self-Balancing Robots: Practical Educational Approach with LabVIEW and Arduino} 
\author[First]{Abdelrahman Abdelgawad} 
\author[First]{Tarek Shohdy} 
\author[First]{Ayman Nada}
\address[First]{Mechatronics and Robotics Engineering, 
    Egypt-Japan University of Science and Technology, Alexandria, Egypt (e-mail: abdelrahman.abdelgawad@ejust.edu.eg, tarek.shohdy@ejust.edu.eg, ayman.nada@ejust.edu.eg)}

\begin{abstract}
A two-wheeled self-balancing robot (TWSBR) is non-linear and unstable system. This study compares the performance of model-based and data-based control strategies for TWSBRs, with an explicit practical educational approach. Model-based control (MBC) algorithms such as Lead-Lag and PID control require a proficient dynamic modeling and mathematical manipulation to drive the linearized equations of motions and develop the appropriate controller. On the other side, data-based control (DBC) methods, like fuzzy control, provide a simpler and quicker approach to designing effective controllers without needing in-depth understanding of the system model. In this paper, the advantages and disadvantages of both MBC and DBC using a TWSBR are illustrated. All controllers were implemented and tested on the OSOYOO self-balancing kit, including an Arduino microcontroller, MPU-6050 sensor, and DC motors. The control law and the user interface are constructed using the LabVIEW-LINX toolkit.  A real-time hardware-in-loop experiment validates the results, highlighting controllers that can be implemented on a cost-effective platform.
\end{abstract}


\end{frontmatter}
%===============================================================================

\section{Introduction}

Contemporary control theory, or model-based control (MBC), originated with
the parametric state-space model by Kalman \cite{Kalman}. MBC entails plant
modeling or identification followed by controller design based on the
assumed accuracy of the plant model, highlighting the crucial role of plant
modeling and identification in MBC theory. Identification theory enables
devising a plant model within a set that accurately represents the true
system or closely approximates it, acknowledging that modeling, whether
through first principles or data-driven identification, entails
approximation and inevitable errors. Unmodeled dynamics are inherent in the
modeling process, making closed-loop control systems designed using MBC
approaches less secure due to these dynamics (\cite{survey}). With
increasing complexity of modern processes, modeling using first principles
or identification has become more challenging, rendering MBC less effective
for modern-day plants. In such scenarios, data-based control (DBC) theory
becomes critical, involving designing controllers directly using
input-output data or data processing knowledge without relying on
mathematical models (\cite{DDC}). In this study, a self-balancing robot was
used as a test bed, akin to an inverted pendulum, maintaining balance by
adjusting motor voltages to control wheel speed (\cite{gonzalez2017low}).
Two-Wheeled Self-Balancing Robots (TWSBRs) are vital for testing control
theories due to their unstable dynamics and nonlinearity, posing challenges
as high-order, multivariable, nonlinear, tightly coupled, and inherently
unstable systems as stated by \cite%
{liang2018differential,act11110339,viswanathan2018integrated}. Balancing
control for such a system is achieved through both linearized and
nonlinear models (\cite{linear-and-non-linear-control, Shaban2014}). Recent studies
have proposed various MBC methods for such systems, including
well-established techniques such as Proportional Integral Derivative (PID)
controllers as in \cite{PID,ren2021control,PID2,Shaban2013} and DBC methods like Fuzzy
Control (\cite{Fuzzy,article}) and Neural Network (NN) control as in \cite%
{NN-control,GANDARILLA2022101259,HOMBURGER20236839}.
These diverse approaches provide options for controlling systems like
self-balancing robots, each with its advantages. This paper conducts a
comprehensive analysis focusing on PID and lead-lag MBC methods and fuzzy
DBC method, aiming to discern their efficacy in regulating self-balancing
robots' behavior. The paper is structured into six sections, starting with
this overview. The subsequent section illustrates dynamic modelling and
state-space representation, followed by examination of simulation framework
and hardware configuration. The fourth section delves into design
methodologies for PID controller, lead-lag compensator, and fuzzy logic
controller, while the fifth section presents a comparative evaluation of
these control strategies. Finally, the paper discusses research results and
future directions.

\section{System Modelling}

\begin{figure}[tb]
\noindent
\centering
\begin{minipage}[tb]{0.5\textwidth}
		\begin{center}
			\includegraphics[width=0.45\textwidth]
			{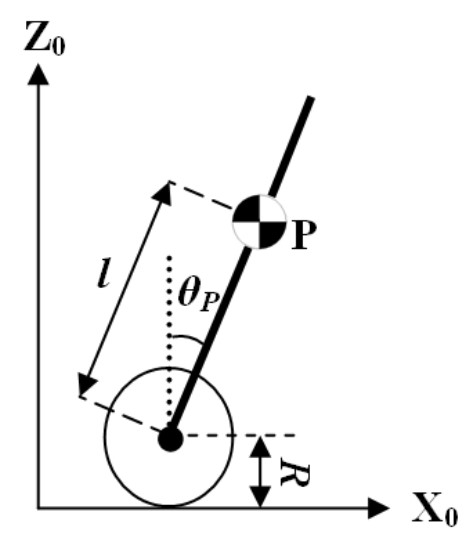}
		\end{center}
	\end{minipage}
\caption{Kinematics of Robot Motion}
\label{fig:Kinematics_of_Robot_Motion}
\end{figure}

In this section, the dynamic model of the TWSBR is derived using Lagrange
equations, as referenced in \cite{dynamic-model}. The notation includes
'trans' for translational and 'rot' for rotational, with '$C$' representing
the chassis and '$W$' for the wheel. A schematic showing the assignment of
state variables is depicted in Fig. \ref{fig:Kinematics_of_Robot_Motion}.
Key parameters include: chassis mass $m,$ wheel mass $M$, wheel's rotational
inertia $J_{w}$, distance from chassis center of mass to wheel center $l$,
wheel radius $R$, gravitational acceleration $g$, wheel-ground friction
coefficient $\mu _{0}$, and and chassis-wheel axis friction coefficient $\mu
_{1}$. $x_{p}$ is the position along $x$-axis, $\theta _{p}$ is the angle of
the chassis inclination to the vertical axis, $z_{p}$ is the position in the 
$z$-axis, $\dot{x}_{p}$ and $\dot{z}_{p}$ are the linear velocities of the
robot. These parameters are identified as $m=0.75\left[ kg\right] $, $M=0.08%
\left[ kg\right] $, $l=0.02$ $\left[ m\right] $, $R=0.035\left[ m\right] $, $%
\mu _{0}=0.1$, and $\mu _{1}=0$. 

\subsection{Kinetic Energy}

\subsubsection{Chassis Kinetic Energy}

The position and velocity of the robot chassis can be expressed as%
\begin{equation*}
\left. 
\begin{array}{c}
x_{p}=x+l\text{ }s\theta _{p} \\ 
z_{p}=l\text{ }c\theta _{p}%
\end{array}%
\right\} \Leftrightarrow \left. 
\begin{array}{c}
\dot{x}_{p}=v+l\omega _{p}\text{ }c\theta _{p} \\ 
\dot{z}_{p}=-l\omega _{p}\text{ }s\theta _{p}%
\end{array}%
\right\}
\end{equation*}

\noindent where $s\theta _{p}=\sin (\theta _{p})$, $c\theta _{p}=cos(\theta
_{p})$, $v=\dot{x}$ is the velocity relative to the ground$,$ and $\omega
_{p}=\dot{\theta}_{p}$.The translating kinetic energy of the chassis can be
expressed as%
\begin{equation}
KE_{trans}^{C}=\frac{1}{2}m[\dot{x}_{p}^{2}+\dot{z}_{p}^{2}]=\frac{1}{2}%
m(v^{2}+l^{2}\omega _{p}^{2})+mvl\omega _{p}c\theta _{p}  \label{eq:KE_new2}
\end{equation}

\noindent The rotational $KE_{rot}^{C}$ of the robot can be expressed as $%
KE_{rot}^{C}=\frac{1}{2}J_{c}\omega _{p}^{2}$, where $J_{c}$ is the
rotational Inertia of the chassis. Adding this term to Equ.\ref{eq:KE_new2},
yields total Kinetic Energy of the chassis ($KE_{total}^{C}$), as%
\begin{equation}
KE_{total}^{C}=\frac{1}{2}m(v^{2}+l^{2}\omega _{p}^{2})+mvl\omega
_{p}c\theta _{p}+\frac{1}{2}J_{c}\omega _{p}^{2}  \label{eq:KE_total-W}
\end{equation}

\subsubsection{Wheels Kinetic Energy}

This is the energy due to the motion of the robot as a whole moving forward
or backward. Each wheel of the robot also contributes to the total kinetic
energy through its rotation. Thus, the kinetic energy of the two wheels ($%
KE_{total}^{W}$) of the robot can be expressed as%
\begin{equation}
KE_{total}^{W}=KE_{trans}^{W}+KE_{rot}^{W}=Mv^{2}+J_{w}\frac{v^{2}}{R^{2}}
\label{eq:KE_total_wheels2B}
\end{equation}

\subsubsection{Total Kinetic Energy of the robot}

The total kinetic energy of the system can be obtained from the combination
of the wheels and chassis kinetic energies as%
\begin{eqnarray}
KE_{total} &=&KE_{total}^{C}+KE_{total}^{W}=\frac{1}{2}mv^{2}+\frac{1}{2}%
ml^{2}\omega _{p}^{2}  \notag \\
&&+mvl\omega _{p}c\theta _{p}+\frac{1}{2}J_{c}\omega _{p}^{2}+Mv^{2}+J_{w}%
\frac{v^{2}}{R^{2}}
\end{eqnarray}

\subsection{Potential Energy and Dissipation Energy}

The only form of potential energy in the system is the gravitational
potential energy, which can be described as%
\begin{equation}
PE_{total}=mglc\theta _{p}  \label{eq:PE}
\end{equation}

\begin{figure}[tb]
\centering
\label{subfig:1ab} \noindent 
\begin{minipage}[tb]{0.25\textwidth}
		\begin{center}
			\includegraphics[width=0.99\textwidth]
			{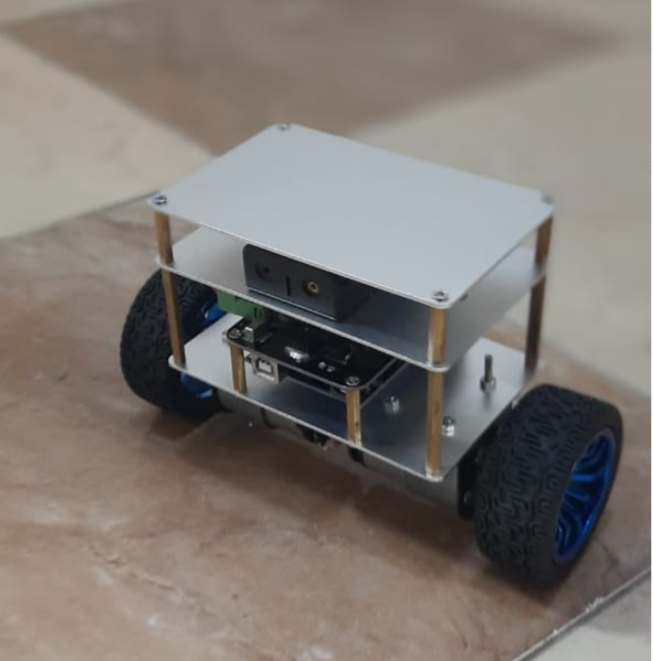}
%			\caption{TWSBR live picture}
%			\label{subfig:live_picture}
		\end{center}
	\end{minipage}
\begin{minipage}[tb]{0.30\textwidth}
	\begin{center}
		\includegraphics[width=0.99\textwidth]
		{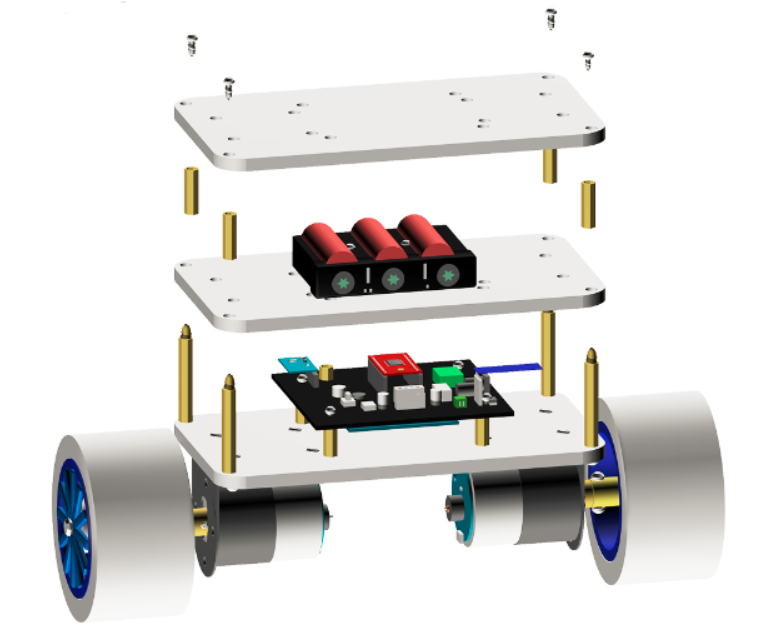}
%		\caption{TWSBR exploded view}
%		\label{subfig:exploded_view}
	\end{center}
\end{minipage}\newline
\caption{TWSBR live and CAD representation}
\end{figure}

\begin{figure*}[tbp]
\noindent 
\begin{minipage}[t]{0.95\textwidth}
		\begin{center}
			\includegraphics[width=1.05\textwidth]
			{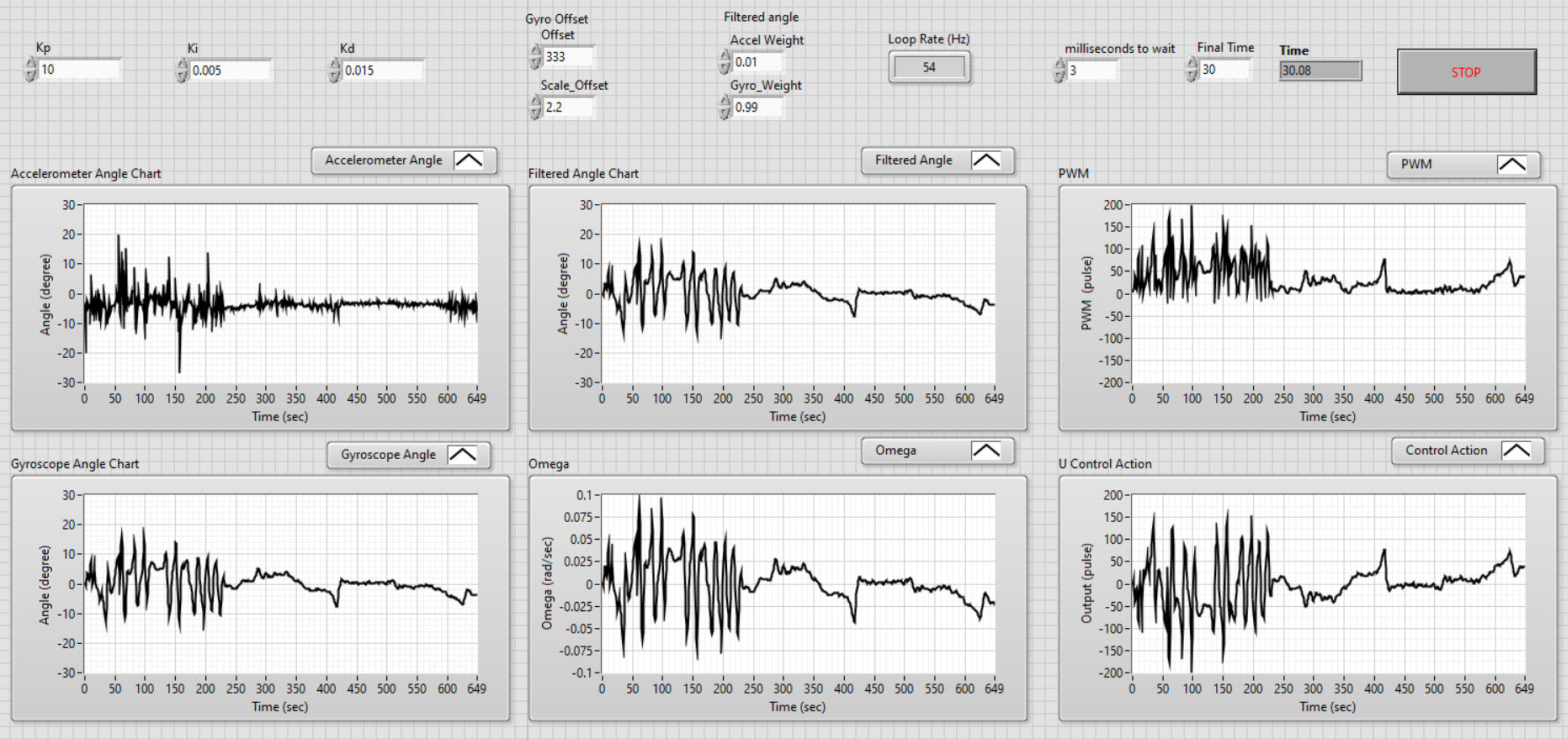}
			\caption{LabVIEW front panel of PID controlled system}
			\label{fig:front panel}
		\end{center}
	\end{minipage}
\end{figure*}

On the other handm the only considerable dissipation that occurs within this
system is the dissipation due to dry friction, which can be described as%
\begin{equation}
F=\mu _{0}v^{2}+\mu _{1}\omega _{p}^{2}  \label{eq:F}
\end{equation}

\noindent where $\mu _{0}$ is the coefficient of friction between the wheels
and the terrain, and $\mu _{1}$ is the coefficient of friction between the
chassis and the wheels rotational axis. By using the Lagrange equation, the
generalized external forces can be described as%
\begin{equation}
{\frac{{\partial }}{{\partial t}}\frac{\partial KE}{\partial \dot{q}_{i}}}-{%
\frac{{\partial KE}}{{\partial q_{i}}}}+{\frac{{\partial F}}{{\partial \dot{q%
}_{i}}}}+{\frac{{\partial PE}}{{\partial q_{i}}}}=Q_{i}
\label{eq:Lagrangian}
\end{equation}

\noindent Here $q_{i}$ is a generalized state variable, and $Q_{i}$ is the
external forces which is the torque exerted by the motors on the wheels. By
substituting in the Lagrange equation (Equ \ref{eq:Lagrangian}), we get the
two differential equations governing the dynamics of the system, Where $\tau
_{L}+\tau _{R}$ are the left and right wheel torques respectively.%
\begin{gather}
(m+2M+\frac{2J_{w}}{R^{2}})\Ddot{x}+mlc\theta _{p}\Ddot{\theta _{p}}+2\mu
_{o}v=0  \label{eq:Lagrangian2} \\
mlc\theta _{p}\Ddot{x}+(ml^{2}+J_{c})\Ddot{\theta _{p}}+2\mu _{1}\omega _{p}
\notag \\
+(mv\omega _{p}-mg)ls\theta _{p}=\tau _{L}+\tau _{R}  \label{eq:Lagrangian3}
\end{gather}

\noindent The Equations of motion, Equ \ref{eq:Lagrangian2} and Equ \ref%
{eq:Lagrangian3}, represent a fully non-linear system. The linearlized form
can be obtained by approximating $\sin \left( \theta _{p}\right) =\theta
_{p} $ assuming $\theta _{p}$ is small and substituting $\cos \left( \theta
_{p}\right) =1$, following the same premise. The linearized dynamic system
equations can be formulated as%
\begin{eqnarray}
(m+2M+2\frac{J_{w}}{R^{2}})\ddot{x}+ml\ddot{\theta _{p}}+2\mu _{o}v &=&0
\label{eq:lin1} \\
ml\Ddot{x}+(ml^{2}+J_{c})\Ddot{\theta _{p}}+(mvlw_{p}-mgl)\theta _{p}+2\mu
_{1}w_{p} &=&\tau _{L}+\tau _{R}  \notag \\
&&  \label{eq:lin3}
\end{eqnarray}

\noindent Defining the state vector as $\mathbf{x}^{T}=%
\begin{bmatrix}
x & \theta _{p} & v & \omega _{p}%
\end{bmatrix}%
$, the state space representation is constructed from the dynamic equations
(Equ \ref{eq:lin1} and \ref{eq:lin3}),%
\begin{equation}
\mathbf{\dot{x}}=\mathbf{Ax}+\mathbf{B}u  \label{eq:mat1}
\end{equation}

\noindent where%
\begin{eqnarray*}
\mathbf{A} &=&%
\begin{bmatrix}
0 & 0 & 1 & 0 \\ 
0 & 0 & 0 & 1 \\ 
0 & -\frac{(ml)^{2}g}{den} & -\frac{2\mu _{o}(ml^{2}+J_{c})}{den} & \frac{%
2ml\mu _{1}}{den} \\ 
0 & \frac{num}{den} & \frac{2\mu _{0}ml}{den} & -\frac{2\mu _{1}\ast num}{den%
}%
\end{bmatrix}
\\
\mathbf{B}^{T} &=&%
\begin{bmatrix}
0 & 0 & -\frac{ml}{den} & \frac{num}{den}%
\end{bmatrix}%
\end{eqnarray*}

\noindent The system output can be described as $y=\mathbf{C}x$, where $%
\mathbf{C}$ as the output matrix. Particularly, $\mathbf{C}=$ $%
\begin{bmatrix}
0 & 1 & 0 & 0%
\end{bmatrix}%
$ if $\theta _{p}$ is the system output. The terms $num=\left( m+2M+2\frac{%
J_{w}}{R^{2}}\right) $, and $den=\left( m+2M+2\frac{J_{w}}{R^{2}}\right)
\left( ml^{2}+J_{c}\right) -(ml)^{2}$.

\section{System and Control Implementation}

\noindent \textbf{Physical Hardware: }In this study, the low cost OSOYOO
TWSBR kit is utilized, which is a tool-set comprising two high-torque and
high-speed gear motors. The TWSBR kit incorporates several components,
including an Arduino UNO Board, Osyoo Balance Robot Shield, MPU6050 Module
equipped with a $3-$axis Accelerometer and $3-$axis Gyroscope, a Bluetooth
module, and a TB6612FNG driver Module. Fig. 2-(a) presents a live depiction
of the TWSBR, while Fig. 2-(b) offers an exploded view from the designed CAD
model, providing a detailed look at the internal components.\newline
\noindent \textbf{Control Design using LabVIEW: }The CAD model was created
in SOLIDWORKS, specifying the mass and inertia properties, with the
simulated mass closely aligning with the actual mass of the robot ($802.45$ $%
grams$ in simulation compared to the measured $805$ $grams$). For the
experimental setup, LabVIEW was utilized to develop the control system for
the TWSBR. The front panel, see Fig. \ref{fig:front panel}, is the interface
for real-time PID control and data analysis. It provides inputs for PID
gains $\left( K_{p},K_{i},K_{d}\right) $ constants for fine tuning based on
the robot's performance. The front panel visualizes the angles measured by
the accelerometer and gyroscope, alongside a filtered angle that combines
these readings for a stable angle measurement using a complementary filter
whose weights are input by user. Additionally, it shows $\omega $, the
angle's rate of change, the control action (saturated between $[-255,255]$),
and the unsigned PWM signals for the motors. The panel includes settings for
experiment length, loop rate, and sensor calibration, offering adaptability
for each test. The PID panel is tailored for its respective control, but the
lead-lag and fuzzy panels only differ in specific gains and parameters, with
a consistent overall design.

\section{Controller Design}

In the controller design process, the Arduino UNO's utilized has high
sampling rate ($16MHz)$, that can develop continuous-time transfer
functions. With the system's sampling rate at $200Hz$, considerably lower
than $16MHz$, we were able to approximate the system to be in analog
platform.

\subsection{PID Controller Design}

\begin{figure}[tb]
\centering
\begin{minipage}[tb]{0.45\textwidth}
    \includegraphics[width=0.9\textwidth]{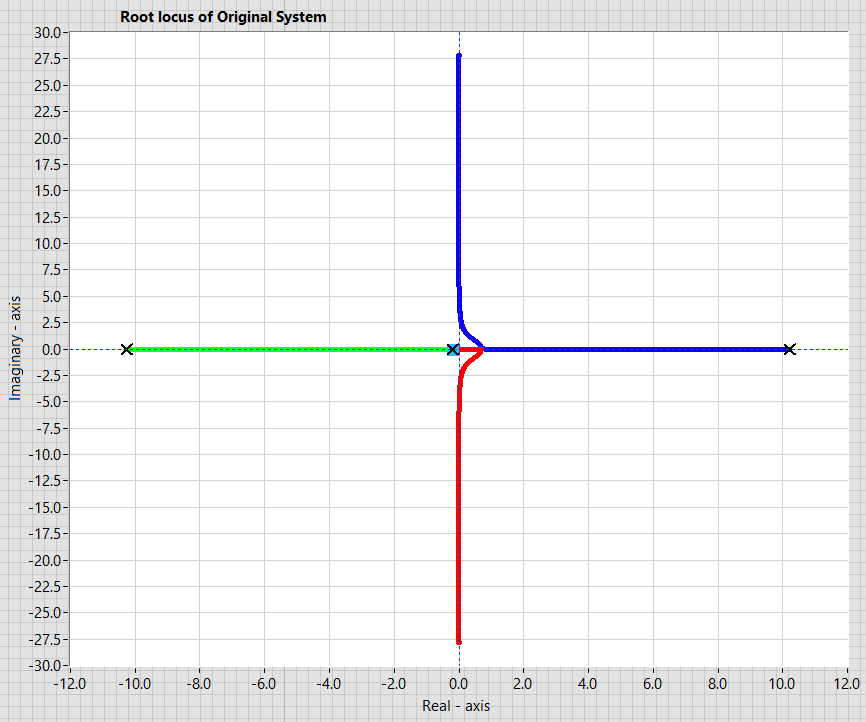}
    \caption{Root locus of the uncompensated system}
    \label{subfig:original_system}
\end{minipage}
\begin{minipage}[tb]{0.45\textwidth}
    \includegraphics[width=0.9\textwidth]{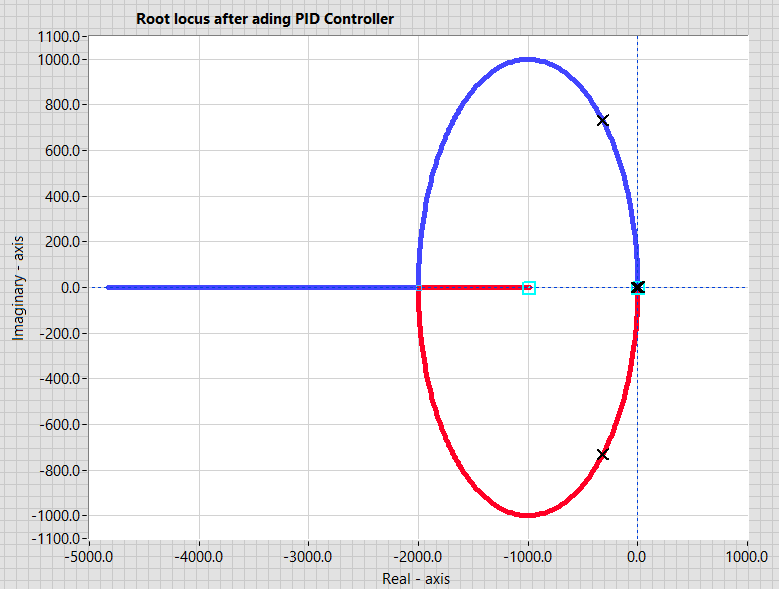}
    \caption{Root locus of PID controlled system: $K_p=10$, $K_i=0.005$, and $K_d=0.015$}
    \label{subfig:pid_refinement}
\end{minipage}
\begin{minipage}[tb]{0.45\textwidth}
    \includegraphics[width=0.9\textwidth]{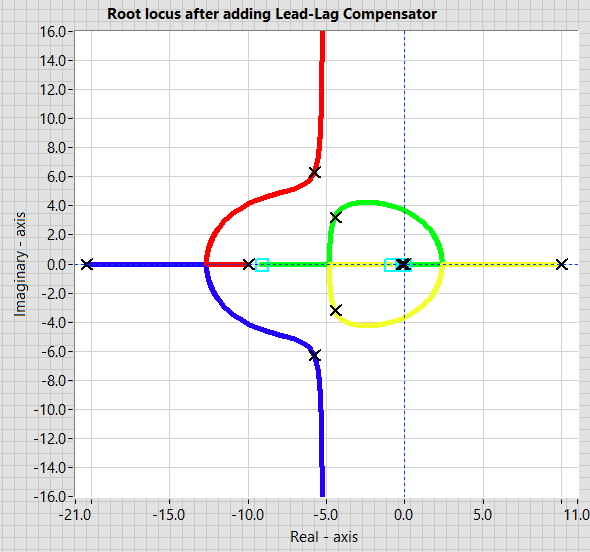}
    \caption{Root locus of Lead-lag compensated system: $K_{c}=3.25$ and $%
G_{c}(s)$ as in Equ.\protect\ref{eq:Lead-Lag}}
    \label{fig:Lead-Lag refinement}
\end{minipage}
\end{figure}

Initially, a PID controller was considered, but the presence of an open-loop
pole on the right-hand side of the $s-$plane made Ziegler-Nichols tuning
impractical, as shown in Fig. \ref{subfig:original_system} which depicts the
root locus of the TWSBR's open-loop system. To overcome this and facilitate
PID controller design, a graphical method using LabVIEW interfacing was
adopted. The design is carried out by locating the zeros of the PD
controller on the root locus to stabilize the system. The steady-state error
required to further add the integral gain to optimize the performance of the
system. The outcomes depicted in Fig. \ref{subfig:pid_refinement} represent
the system's response with applying specific gains determined through
graphical analysis, resulting in a stable operating region. 

\subsection{Design of Lead-lag Compensator}

The lead-lag compensator enhances feedback control system performance by
simultaneously addressing stability, transient response, and steady-state
accuracy. Its primary function lies in introducing both phase lead and phase
lag within the system's frequency response. This dual-phase adjustment
bolsters system stability by augmenting the phase margin, fortifying
resilience against oscillations and instability. The phase lead improves
transient response, facilitating accelerated response times, diminishing
settling time, and mitigating oscillations for elevated dynamic performance.
On the other hand, the phase lag refines the steady-state accuracy by
attenuating higher-frequency components and adjusting the gain at lower
frequencies, minimizing steady-state error, particularly in scenarios with
disturbances. However, designing and deploying a lead-lag compensator
requires careful consideration of inherent trade-offs in balancing
stability, transient response, and steady-state accuracy. Achieving an
optimal configuration involves analysis of system dynamics and performance
requirements, often requiring sophisticated mathematical modeling and
extensive tuning efforts. The system achieves stability through a lead-lag
compensator, see Fig. \ref{fig:Lead-Lag refinement}, which displays the
open-loop root locus of the compensated system. Design requirements included
a settling time of $0.7[s]$ and a $6\%$ overshoot, achieved by positioning
the poles and zeros on the root locus using the controller in Equ. \ref%
{eq:Lead-Lag}. Specifically, the lead controller's time constant is set to $%
0.1095[s]$ with an $\alpha =0.4494$, while the lag controller's time
constant is $1.123[s]$ with a $\beta =7.1439$ \ Additionally, the lead-lag
controller gain $K_{c}$ is set to $3.25$. 
% \begin{figure}[tb]
% \noindent 
% \begin{minipage}[tb]{0.50\textwidth}
% 		\begin{center}
% 			\includegraphics[width=0.90\textwidth]
% 			{images/Lead Lag refinement.png}
% 		\end{center}
% 	\end{minipage}\newline
% \caption{Root locus of Lead-lag compensated system: $K_{c}=3.25$ and $%
% G_{c}(s)$ as in Equ.\protect\ref{eq:Lead-Lag}}
% \label{fig:Lead-Lag refinement}
% \end{figure}

\begin{equation}  \label{eq:Lead-Lag}
G_{c}^{lead-lag}(s)= K_c \frac{0.13998s^2 + 1.40322s + 1.1381}{s^2 +
20.4484s + 2.53227}
\end{equation}

\subsection{Fuzzy Logic Controller (FLC)}

The design process of a fuzzy logic controller involves three crucial
stages: Fuzzification, where crisp inputs are transformed into fuzzy sets to
handle linguistic variables and uncertainties in real-world systems; Fuzzy
rule-based decision making, using predefined rules to guide the controller's
actions based on fuzzy inputs through complex decision spaces; and
Defuzzification, where fuzzy control actions are converted into crisp values
for practical implementation using methods such as centroid, mean of maximum
(MOM), or weighted average.

By traversing through these stages, the FLC can process fuzzy inputs and
generate precise control actions to steer complex systems towards desired
outcomes. A notable implementation of FLC is the PID-like FLC, which
combines proportional and derivative control elements \cite{fuzzy-ayman-nada}%
. This approach is advantageous in real-time applications on nonlinear
mechatronic systems where traditional Model-Based Controllers may struggle.
The PID-like FLC thus offers a versatile and efficient solution for
controlling complex systems. In the practical implementation phase, we
integrated the PID-like Fuzzy Logic Controller (FLC) as outlined by \cite%
{fuzzy-ayman-nada}. Transition to real-world applications was streamlined by
easily adapting control parameters: PID Proportional Gain $K_{p}$ determines
the error term's proportional contribution to the control output; The
Derivative Gain $K_{d}$ accounts for the contribution of the rate of change
of the error term over time; Integral Gain $K_{i}$: compensates for
accumulated past errors, aiding in the elimination of steady-state error; \
and the FLC Proportional Gain $K_{u}$ influences the impact of fuzzy rules
on the control action. In the implementation phase, we fine tuned the three
gains explained before. We reached the balance using $K_{p}=150$, $K_{i}=1.5$%
, $K_{d}=1$, $K_{u}=1$, based on a trial and error approach which led to
acceptable performance.

\section{Results and Discussion}

\subsubsection{Experimental Validation}

\begin{figure}[tb]
\begin{minipage}[tb]{0.45\textwidth}
    \centering
    \includegraphics[width=0.9\textwidth]{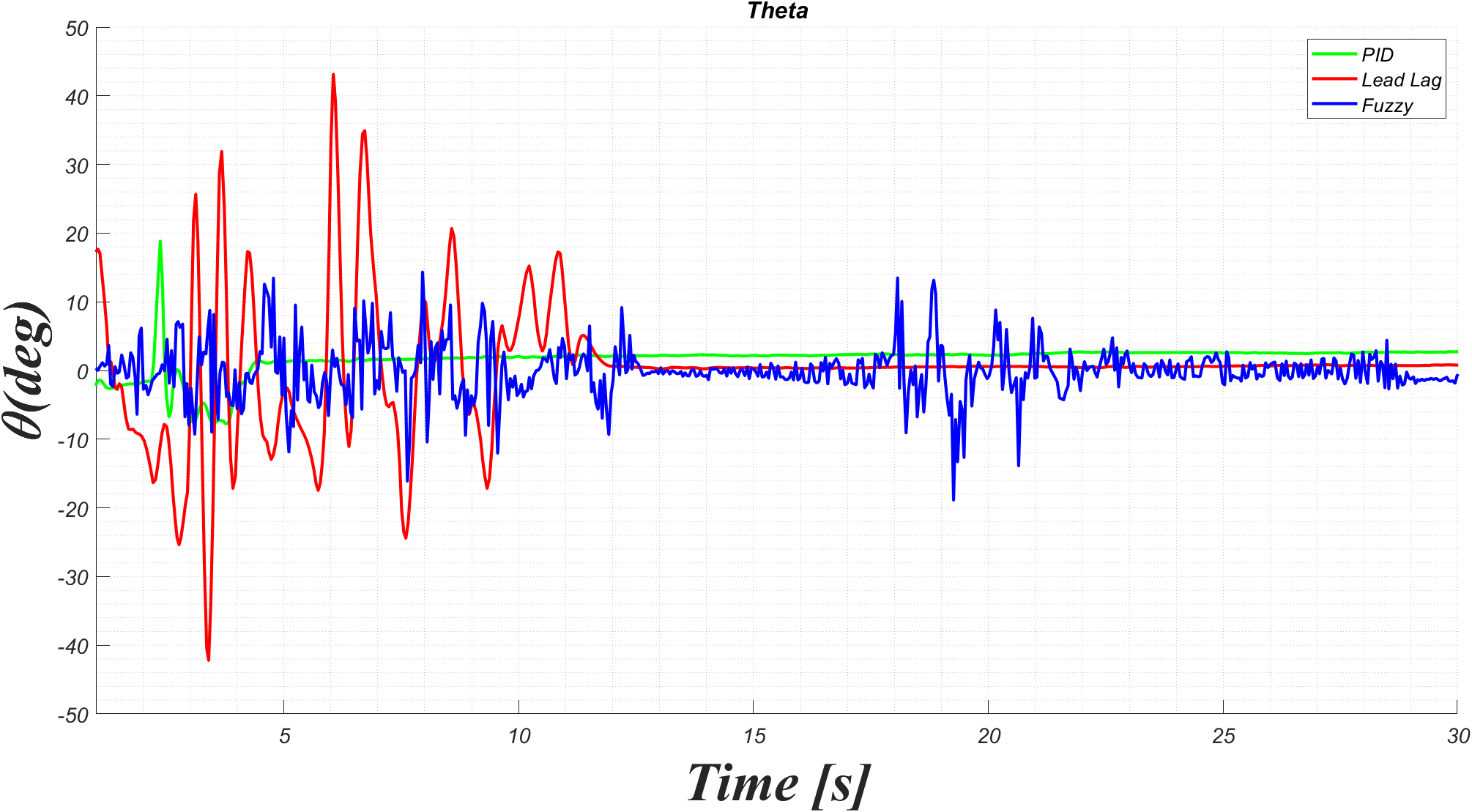}
    \caption{Effect of PID, lead lag, and FLC on $\theta_p$}
    \label{fig:Controllers_Theta}
\end{minipage}%
\hspace{0.05\textwidth}
\begin{minipage}[tb]{0.45\textwidth}
    \centering
    \includegraphics[width=0.9\textwidth]{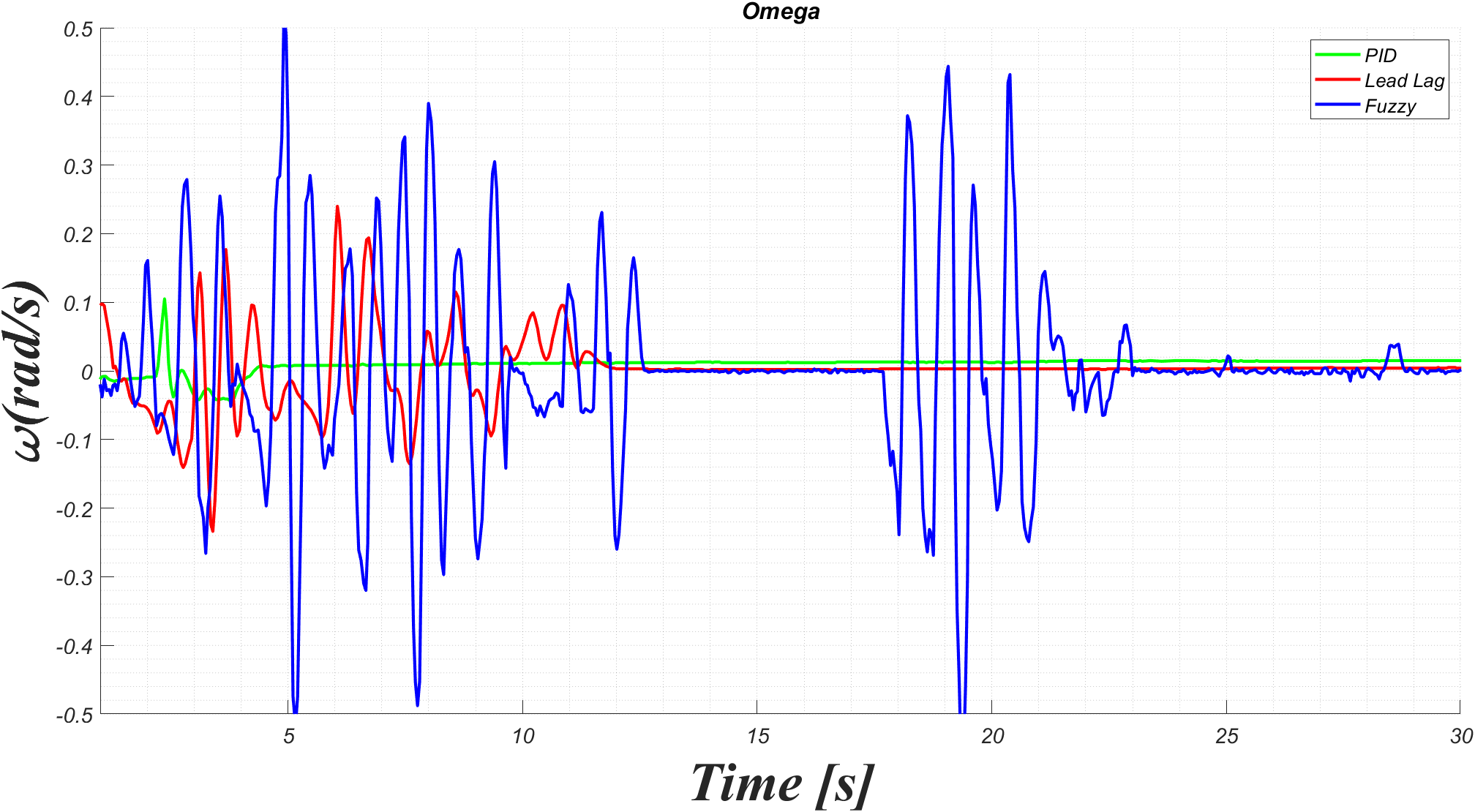}
    \caption{Effect of PID, lead lag, and FLC on $\omega$}
    \label{fig:Controllers_Omega}
\end{minipage}

\begin{minipage}[tb]{0.45\textwidth}
    \centering
    \includegraphics[width=0.9\textwidth]{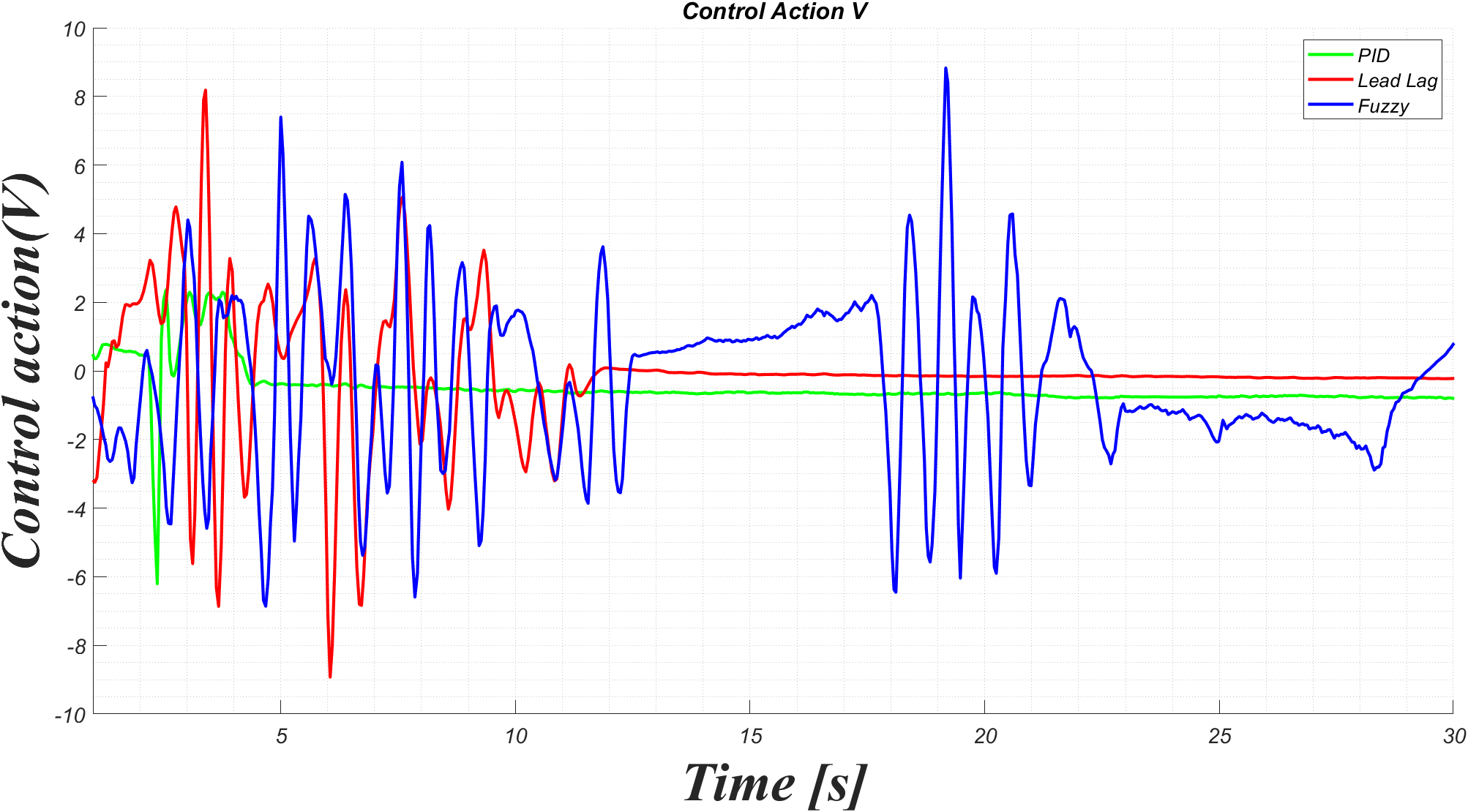}
    \caption{Effect of PID, lead lag, and FLC on control action}
    \label{fig:Controllers_Control_Action}
\end{minipage}%
\hspace{0.05\textwidth}
\begin{minipage}[tb]{0.45\textwidth}
    \centering
    \includegraphics[width=0.9\textwidth]{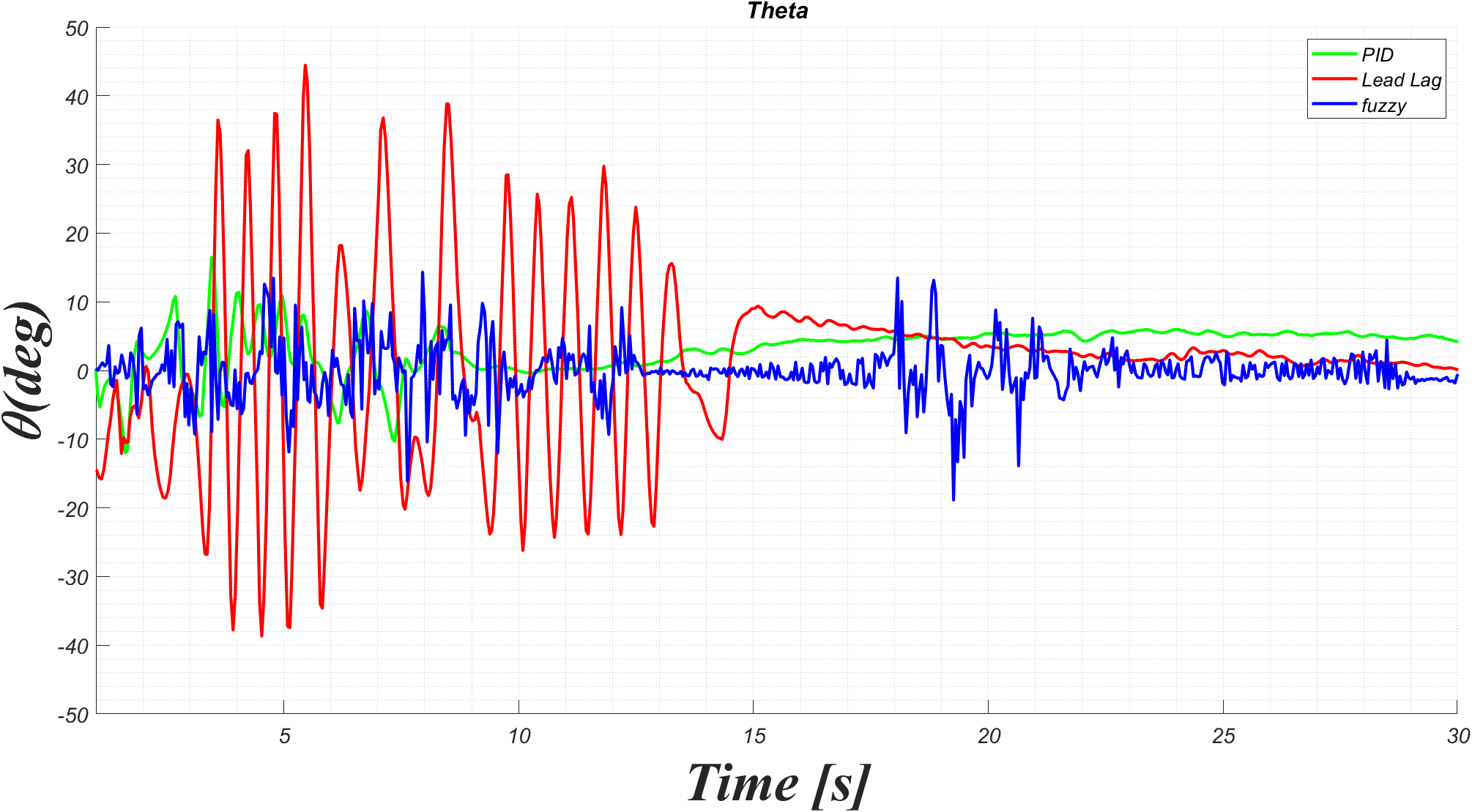}
    \caption{Effect of controllers on $\theta$ with mass uncertainty}
    \label{fig:Controllers_Theta_with_mass}
\end{minipage}

\begin{minipage}[tb]{0.45\textwidth}
    \centering
    \includegraphics[width=0.9\textwidth]{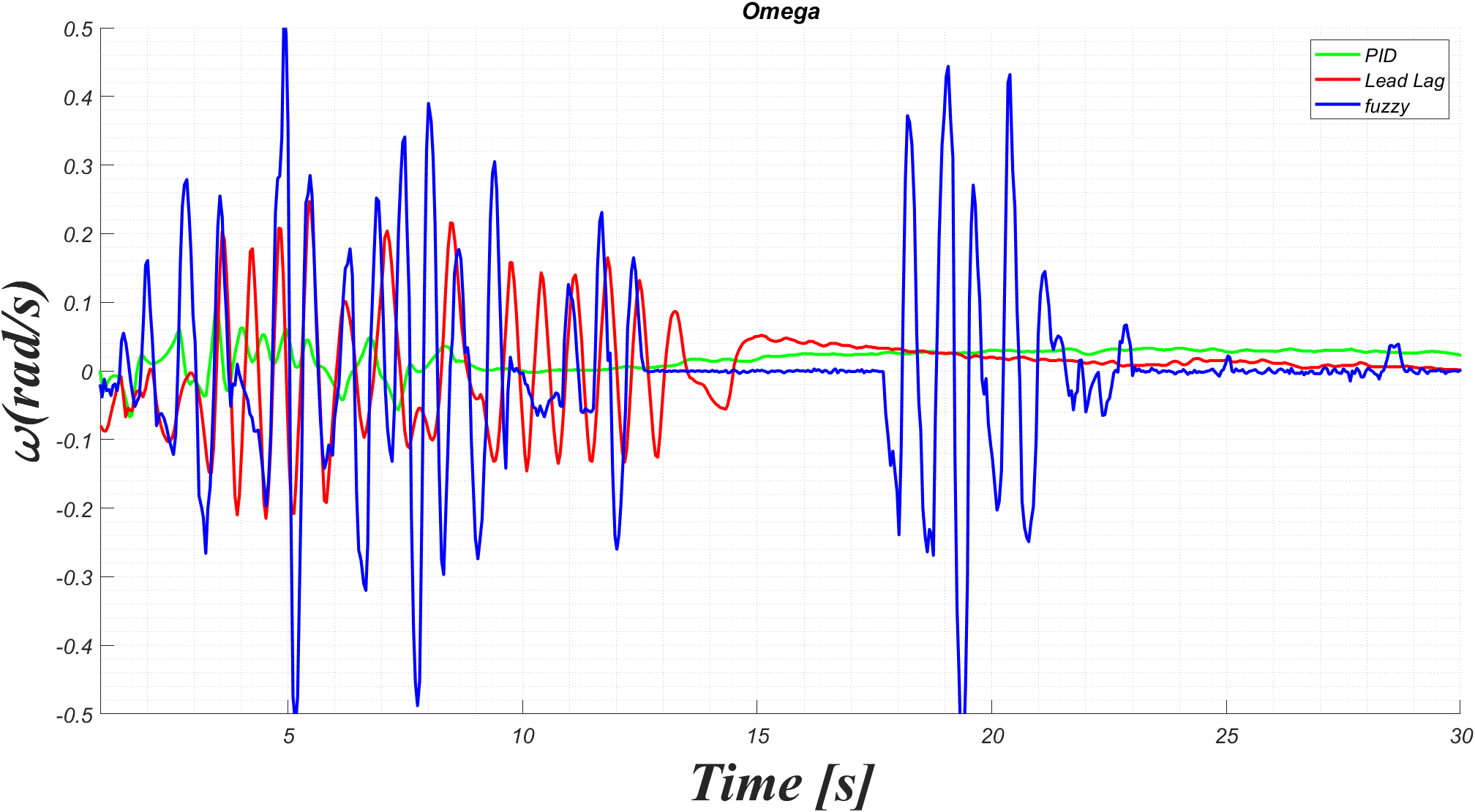}
    \caption{Effect of controllers on $\omega$ with mass uncertainty}
    \label{fig:Controllers_Omega_with_mass}
\end{minipage}%
\hspace{0.05\textwidth}
\begin{minipage}[tb]{0.45\textwidth}
    \centering
    \includegraphics[width=0.9\textwidth]{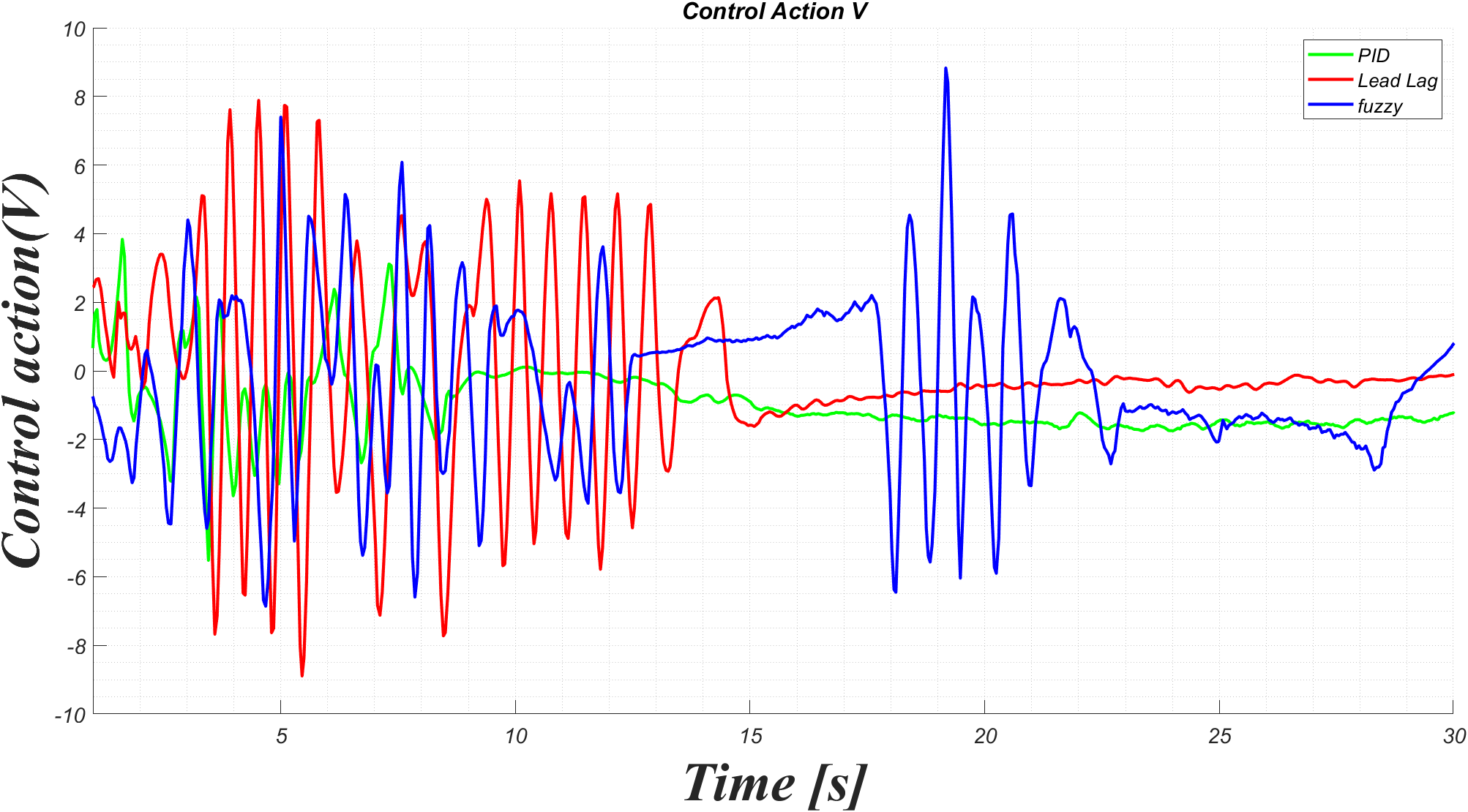}
    \caption{Effect of controllers on control action with mass uncertainty}
    \label{fig:Controllers_Controlled_Action_with_mass}
\end{minipage}
\end{figure}

For a rigorous comparative analysis of the controllers' performance, a $30[s]
$ test duration was selected. This interval facilitated evaluating each
controller's impact on the tilting angle $\theta _{p}$, its rate of change ($%
\omega $), and motor control voltage, see Fig. \ref{fig:Controllers_Theta},
Fig. \ref{fig:Controllers_Omega}, and Fig. \ref%
{fig:Controllers_Control_Action}, respectively. To conduct this comparison
and test the performance of each controller, we ensured consistency by using
identical components and testing conditions throughout all experiments. We
tested the robustness by introducing uncertainties to the system parameters
where we added an additional mass to the platform and tested it with the
same controller design and tuned gains, as shown in Fig. \ref%
{fig:Controllers_Theta_with_mass}, Fig. \ref{fig:Controllers_Omega_with_mass}%
, and Fig. \ref{fig:Controllers_Controlled_Action_with_mass}. 

\subsubsection{Evaluation Metrics}

\begin{enumerate}
\item \textbf{Steady-State Accuracy}: MBC classical approaches (PID - lead
lag) are the most efficient for regulating the tilt angle, see Fig. \ref%
{fig:Controllers_Theta} and their exclusive behaviour on the change of $%
\omega $ as shown in Fig. \ref{fig:Controllers_Omega} with PID being
slightly better.

\item \textbf{Settling Time}: Regarding settling time, MBC methods
outperform, with PID and lead-lag controllers reaching settling times of $5$
and $12[s]$, respectively. In contrast, the DBC strategy, specifically the
FLC, attains a settling time of $27[s]$, as depicted in Fig. \ref%
{fig:Controllers_Theta}.

\item \textbf{Ease of Design}: FLC does not require a predefined
mathematical model, simplifying its design compared to MBC approaches.
However, the aggressive control action of FLC, as shown in Fig. \ref%
{fig:Controllers_Control_Action}, may not be cost-efficient. Nonetheless,
FLC is ideal for quick model-free implementations.

\item \textbf{Tuning Complexity}:Regarding the ease of design, PID and
lead-lag controllers allow for straightforward implementation once detailed
design is complete. Conversely, although Fuzzy Logic Controllers (FLC) have
a relatively simple design process, their tuning is complex and
time-consuming.

\item \textbf{Robustness}: When adding an additional mass to induce
uncertainties in the model to test robustness, All of PID, lead lag and FLC
controllers could maintain the platform stable in Fig. \ref%
{fig:Controllers_Theta_with_mass}, Fig. \ref{fig:Controllers_Omega_with_mass}%
, and Fig. \ref{fig:Controllers_Controlled_Action_with_mass}. The PID
controller had the best performance having the lowest overshoot and settling
time as shown in Fig. \ref{fig:Controllers_Theta_with_mass}, with less
aggressive control actions as in Fig. \ref%
{fig:Controllers_Controlled_Action_with_mass} which makes the angle rate of
change low, see Fig. \ref{fig:Controllers_Omega_with_mass}. We can see that
the FLC was not able to reach the steady state within the 30 seconds trial
while also having the greatest overshoot.
\end{enumerate}

% With a deep observation on the performance of the PID controller,

\section{Conclusion}

This study presents a comparative analysis between Model-Based Control (MBC)
and Data-Based Control (DBC) strategies, specifically the PID, lead-lag, and
fuzzy logic for stabilizing a Two-Wheeled Self-Balancing Robot. A
comprehensive control framework is established, including the state-space
model, LabVIEW interface, and real-time prototyping of the robot. The
analysis highlights the the distinct performance characteristics of each
method. MBC methods, particularly PID and lead-lag controls, show better
accuracy, minimum error, faster response, and high robustness, making them
suitable for precise control requirements. Despite the ease of design with
fuzzy logic controllers (FLC), they tend to have more fluctuated control
actions and need a relatively complicated tuning procedure. It is suggested
that FLC's performance could be optimized by specific tuning of its
membership functions and rules. The introduced framework throughout the
paper enables young engineers to realize the control system design, and
improve their skills toward system optimization and real-time implementation.

\bibliographystyle{plain}
\bibliography{references}

\end{document}